\documentclass[conference]{IEEEtran}
\IEEEoverridecommandlockouts

\usepackage{cite}
\usepackage{amsmath,amssymb,amsfonts}
\usepackage{algorithmic}
\usepackage{graphicx}
\usepackage{textcomp}
\usepackage{xcolor}
\usepackage{amsthm}
\usepackage{comment}
\usepackage{url}

\def\BibTeX{{\rm B\kern-.05em{\sc i\kern-.025em b}\kern-.08em
    T\kern-.1667em\lower.7ex\hbox{E}\kern-.125emX}}
\makeatletter
\newcommand{\linebreakand}{%
  \end{@IEEEauthorhalign}
  \hfill\mbox{}\par
  \mbox{}\hfill\begin{@IEEEauthorhalign}
}
\makeatother

\theoremstyle{definition}

\begin{document}

\title{Graphint: Graph-based Time Series Clustering Visualisation Tool}

\author{
\IEEEauthorblockN{Paul Boniol}
\IEEEauthorblockA{\textit{Inria, ENS, PSL, CNRS} \\
paul.boniol@inria.fr}
\and
\IEEEauthorblockN{Donato Tiano}
\IEEEauthorblockA{\textit{Università di Modena} \\
donato.tiano@unimore.it}
\and
\IEEEauthorblockN{Angela Bonifati}
\IEEEauthorblockA{\textit{Lyon 1 University, IUF, Liris CNRS} \\
angela.bonifati@univ-lyon1.fr}
\and
\IEEEauthorblockN{Themis Palpanas}
\IEEEauthorblockA{\textit{Université Paris Cité - IUF} \\
themis@mi.parisdescartes.fr}
}

\maketitle

\begin{abstract}
With the exponential growth of time series data across diverse domains, there is a pressing need for effective analysis tools. Time series clustering is important for identifying patterns in these datasets. However, prevailing methods often encounter obstacles in maintaining data relationships and ensuring interpretability. We present Graphint, an innovative system based on the $k$-Graph methodology that addresses these challenges. Graphint integrates a robust time series clustering algorithm with an interactive tool for comparison and interpretation. More precisely, our system allows users to compare results against competing approaches, identify discriminative subsequences within specified datasets, and visualize the critical information utilized by $k$-Graph to generate outputs. Overall, Graphint offers a comprehensive solution for extracting actionable insights from complex temporal datasets.
\end{abstract}


\maketitle

\section{Introduction}

We are witnessing a surge in the production of time series data, which necessitates novel solutions for effectively analyzing these vast datasets. 
This need is particularly pressing for applications across diverse domains~\cite{Palpanas2019}.
Time series clustering~\cite{aghabozorgi2015time, wang2006characteristic,vizClustdemo,6065026} is crucial for uncovering patterns within time series data. 
The primary goal is to partition a dataset of $N$ time series into $k$ predetermined clusters.
Various approaches address this challenge.
Raw-based methods, such as $k$-Means and $k$-Shape~\cite{paparrizos2015k}, operate directly on raw data, preserving information and temporal dependencies. While effective, these methods can struggle with noise, which can obscure meaningful patterns. 
Moreover, clustering raw time series might result in clusters that lack clear distinctions or meaningful insights.

Feature-based approaches, such as FeatTS~\cite{tiano2021featts} or Time2Feat~\cite{bonifati2022time2feat}, tackle the challenges associated with raw-based methods. These approaches often involve dimensionality reduction, which enhances discrimination and offers interpretability. 
FeatTS and Time2Feat utilize feature extraction techniques to cluster both univariate and multivariate time series but they may experience information loss due to transforming sequential patterns into features and face difficulties in managing high-dimensional feature spaces.

\begin{figure}
    \centering
    \includegraphics[width=1\linewidth]{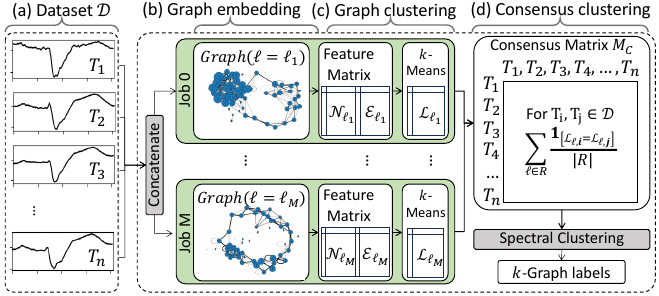}
    \vspace{-0.6cm}
    \caption{$k$-Graph Pipeline~\cite{kGraph}.}
    \vspace{-0.4cm}
    \label{fig:introfig}
\end{figure}

Deep learning approaches, exemplified by Deep Auto-Encoder (DAE)~\cite{tian2014learning} and Deep Temporal Clustering (DTC)~\cite{DTCAlgorithm}, leverage deep learning for superior accuracy. 
DAE transforms raw input data into novel space representations, facilitating seamless integration into clustering algorithms. 
DTC uses DAE for initialization, leveraging learned features for indirect clustering. 
Despite their high efficacy, deep learning approaches face challenges related to the interpretability of their decision-making processes and may struggle to incorporate domain-specific knowledge.

This article presents Graphint\footnote{\url{https://graphit.streamlit.app/}, \url{https://github.com/boniolp/graphit}}, a system designed to address these limitations. 
Its objectives are twofold: (i) to provide an effective time series clustering algorithm, and (ii) to offer an interactive tool that facilitates the interpretation and comparison of time series clustering results.
This system is based on $k$-Graph~\cite{kGraph}, a novel and interpretable graph-based method for time series clustering. Figure~\ref{fig:introfig} illustrates the three steps of the pipeline: the \textbf{Graph Embedding} phase, where a set of graphs is computed based on subsequences of varying lengths from the time series dataset. The \textbf{Graph Clustering} stage follows, where features are extracted from the time series using information embedded in the graph nodes and edges, and then clustered with the $k$-Means algorithm. In the \textbf{Consensus Clustering} phase, spectral clustering establishes consensus across multiple partitions obtained in the previous step. To enhance the interpretability of the system, $k$-Graph introduces an additional step known as \textbf{Interpretability Computation}. This step aims to identify the most relevant subgraph, thereby improving the comprehensibility of the system.

\section{The $k$-Graph Approach}

\label{sec:kGraph_approach}
\textbf{[Background]}: A time series $T$ in $\mathbb{R}^n$ represents a sequence of real numbers, where each $T_i$ denotes a point in the series. Subsequences, such as $T_{i,\ell}$, are local segments of interest within $T$, comprising $\ell$ consecutive points starting from index $i$.
Defined with a Node Set $\mathcal{N}$ consisting of unique integers, and an Edge Set $\mathcal{E}$ containing tuples $(x_i,x_j)$, where $x_i, x_j \in \mathcal{N}$, graphs are represented as ordered pairs $\mathcal{G}=(\mathcal{N},\mathcal{E})$. $k$-Graph exclusively employs directed graphs, denoted as $\mathcal{G}$, where $\mathcal{E}$ represents an ordered collection of node pairs within $\mathcal{N}$.
Time series clustering is fundamental for identifying meaningful patterns and aims to group similar time series.  
The challenge of time series clustering involves partitioning a dataset of $n$ time series, denoted as $D$, into $k$ clusters $(C = {C_{1}, C_{2}, ..., C_{k}})$, with $k$ the number of clusters.

\noindent\textbf{[Problem Formulation]}: $k$-Graph presents a new approach to time series clustering using a graph embedding. 
It transforms the time series dataset into a sequence of abstract states represented as nodes in a directed graph $\mathcal{G}=(\mathcal{N},\mathcal{E})$, where $\mathcal{N}$ denotes nodes and $\mathcal{E}$ represents edges capturing transition frequency between states.
Nodes in this graph correspond to similar patterns across various time series, while edges represent transitions between these patterns. 
Consequently, similar time series are grouped within the same region of the graph: in a dataset $\mathcal{D} = \{T_0, T_1, ..., T_n\}$, a specific cluster of time series, $C_i \subset \mathcal{D}$, corresponds to a distinguishable subgraph $\mathcal{G}_{C_i} \subset \mathcal{G}$ called a \emph{Graphoid}.
By definition, the Graphoid of a cluster includes all nodes and edges traversed by at least one time series of that cluster. 
Thus, a node in $\mathcal{G}$ may belong to multiple Graphoids.
This approach treats all nodes equally, without distinguishing between those representing patterns across all clusters (representativity) and those exclusive to specific clusters (exclusivity): each node may capture both common and unique patterns across different clusters. 

The $k$-Graph approach introduces the notions of representativity and exclusivity of nodes and edges. 
The representativity of a node in a cluster $C_i$ ($|N|_{C_i}$), indicates the proportion of time series within a cluster that passes through the node, divided by the total number of time series in that cluster. Conversely, the exclusivity of a node in a cluster $C_i$ ($Pr_{C_i}(N)$), represents the proportion of time series within a cluster that traverse the node, divided by the total number of time series traversing the same node. Similar definitions apply to edges.
These properties enable us to refine the concept of Graphoid into two types:
(i) the $\lambda$-$Graphoid$ comprises nodes and edges sufficiently representative of a cluster (i.e., at least $\lambda$\% of the time series of a given cluster cross these nodes and edges), where $\lambda$ defines the representativity threshold; 
(ii) the $\gamma$-$Graphoid$ comprises nodes and edges that occur exclusively in the corresponding cluster (i.e., at least $\gamma$\% of the time series crossing the edges and the nodes belong to this cluster only), with the frequency threshold determined by $\gamma$. 
For example, a node belongs to the $\lambda$-Graphoid of a cluster with $\lambda=0.8$, if $80$\% of the time series of $C$ are crossing that node. 
Moreover, a node belongs to the $\gamma$-Graphoid of a cluster with $\gamma=0.8$ if $80$\% of the time series crossing that node belongs to this cluster.
These definitions facilitate the extraction of subgraphs, highlighting a cluster's most significant nodes and edges and, thus, its unique temporal patterns.

The perfect scenario is characterized by each cluster having its own unique set of nodes and edges in the graph, without any overlap between clusters. This implies that the patterns specific to each cluster are exclusively represented within its designated graph portion. 
However, achieving such a clear-cut partitioning is not always possible. 
Instead, the focus shifts to ensuring each cluster contains nodes and edges that effectively capture its distinctive patterns. 
These representative and exclusive nodes and edges should be prevalent within the cluster while rare in others.

\subsection{$k$-Graph Computation Steps}

We recap the $k$-Graph time series clustering solution, which involves three main steps as depicted in Figure \ref{fig:introfig}.

The first step of $k$-Graph is \textbf{Graph Embedding}, where the goal is to create a single graph representing the entire time series dataset for clustering (inspired by Series2Graph~\cite{Series2GraphPaper}). 
In practice, for a dataset $\mathcal{D}$, $k$-Graph constructs $M$ graphs $\mathcal{G}_\ell=(\mathcal{N}_\ell,\mathcal{E}_\ell)$ for $M$ subsequence lengths $\ell$, shown in Figure \ref{fig:introfig}(b). 
For each graph, Principal Component Analysis (PCA) is applied, allowing us to project the subsequences into a two-dimensional space while retaining their essential shapes. Nodes representing dense regions are generated via local maxima identification using radial scan and kernel density estimation. Finally, $k$-Graph connects nodes to capture transitions between patterns by establishing edges corresponding to sequential subsequences within time series.

In the \textbf{Graph Clustering} phase (Figure \ref{fig:introfig}(c)), $k$-Graph utilizes the $M$ distinct graphs obtained from $M$ different subsequence lengths to cluster dataset $\mathcal{D}$. For each time series, two types of features are generated: node-based and edge-based, by counting intersections with nodes and edges in the graph. In the resulting feature matrix, rows and columns represent time series and nodes or edges. 
The clustering algorithm $k$-Means is then applied to group time series based on these features, producing a partition $\mathcal{L}_\ell$ for each graph $\mathcal{G}_\ell$.

At this stage, $k$-Graph generates one clustering partition $\mathcal{L}_\ell$ for each graph produced in the graph embedding phase, resulting in a total of $M$ partitions. The objective now is to synthesize a consensus from these partitions. $k$-Graph constructs a consensus matrix, $M_C$, during the \textbf{Consensus Clustering} step. This matrix measures the frequency with which time series pairs are grouped in the same cluster across different graphs, as shown in Figure\ref{fig:introfig}(d). 
We finally apply spectral clustering on this matrix and produce a final clustering partition $\mathcal{L}$, corresponding to the final $k$-Graph labels.

\subsection{Graph Interpretability}

$k$-Graph operates with multiple graphs (denoted as $\mathcal{G}_\ell$ for each subsequence length $\ell$) to generate a partition. The goal is to identify the most representative graph from this ensemble.
To achieve this, $k$-Graph employs the following two criteria:

\noindent \textbf{[Consistency]:} To assess the consistency between the final labels $\mathcal{L}$ and those produced by each graph $\mathcal{G}_\ell$, $k$-Graph use the Adjusted Rand Index (ARI). This criterion, noted $W_c(\ell) = ARI(\mathcal{L},\mathcal{L}_\ell)$, quantifies the agreement between $k$-Graph labels and those associated with a specific subsequence length $\ell$.

\noindent \textbf{[Interpretability Factor]:} The second criterion selects the most interpretable graphs and a given clustering partition. Interpretability is achieved by ensuring each cluster has a unique set of nodes and edges within the graph without overlapping with other clusters. $k$-Graph calculates the interpretability factor $W_e(\ell)$ for $\mathcal{G}_\ell$ by averaging the maximum exclusivity of nodes across all clusters.

Finally, to select the optimal graph length $\bar{\ell}$, $k$-Graph chooses the $\ell$ value from the set $R$ that maximizes the product of the interpretability factor $W_e(\ell)$ and consistency $W_c(\ell)$. Subsequently, the graph associated with this length is used to compute the graphoids.
In computing them, $k$-Graph evaluates the representativity and exclusivity of each node in $\mathcal{G}_\ell$. Nodes are selected to meet the criteria for $\lambda$ and $\gamma$-Graphoid based on desired values of $\lambda$ and $\gamma$.

\section{Graphint: System Overview}
\begin{figure}
    \centering
    \includegraphics[width=1\linewidth]{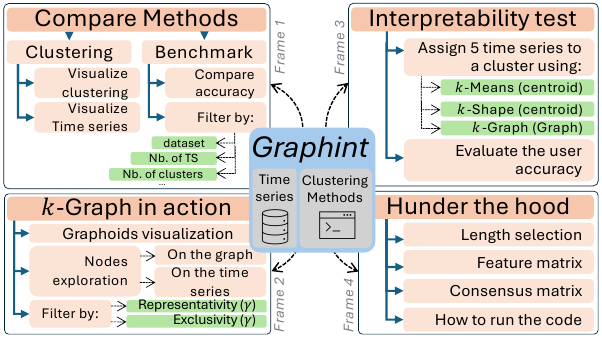}
    \vspace*{-0.8cm}
    \caption{Graphint system overview.}
    \vspace*{-0.4cm}
    \label{fig:system}
\end{figure}

\begin{figure*}
    \centering
    \includegraphics[width=1\linewidth]{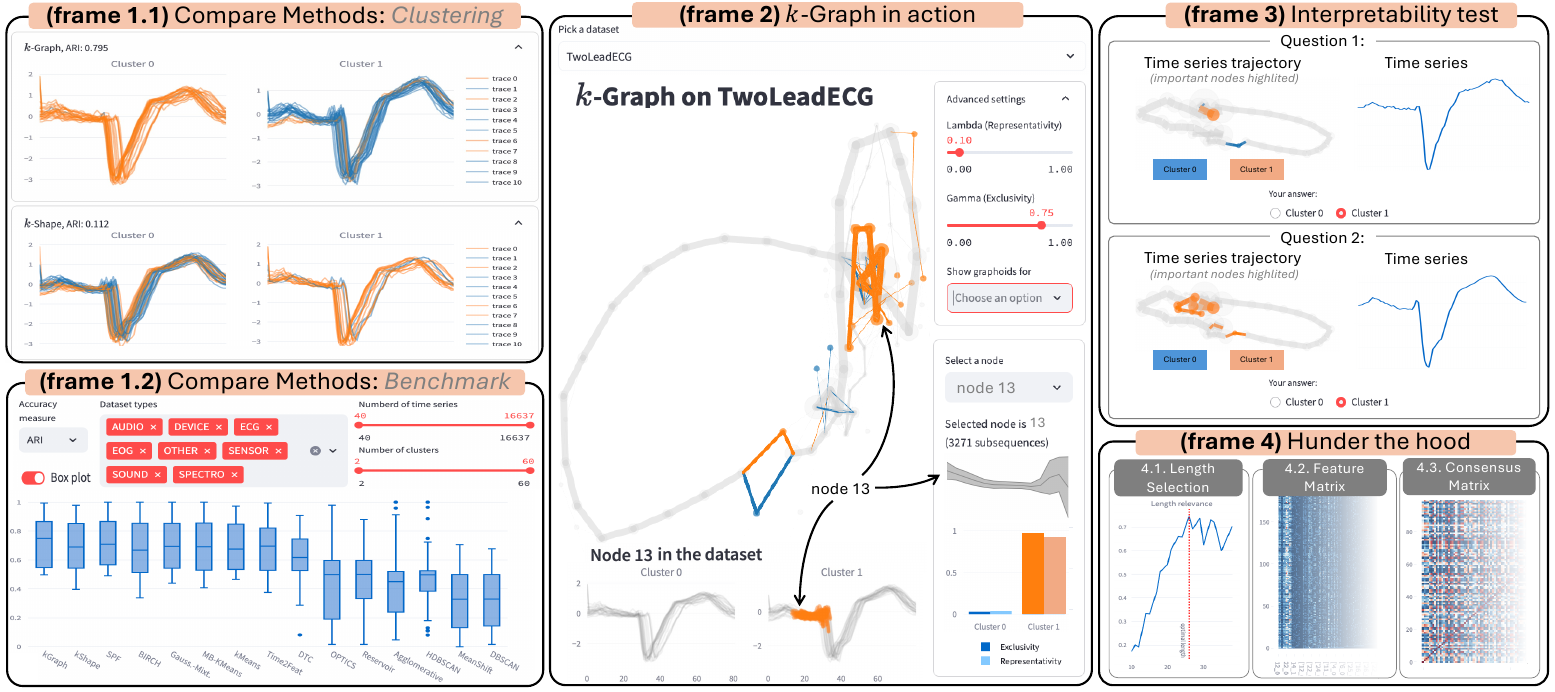}
    \vspace*{-0.8cm}
    \caption{Main frames of Graphint.}
    \vspace*{-0.4cm}
    \label{fig:frames}
\end{figure*}

In this section, we describe Graphint, the system designed to help users and analysts utilize $k$-Graph and interpret clustering results.
The GUI is a stand-alone web application developed using Python 3.9 and the Streamlit Framework.
Figure \ref{fig:system}
illustrates Graphint's functionalities. 
The system accesses a collection of datasets, as well as a set of time series clustering methods. 
The user selects a dataset and receives from GraphHint (i) the possibility to compare time series clustering methods, (ii) interpret the clustering of $k$-Graph, (iii) Compare the interpretability power of $k$-Graph with two baselines, or (iv) dive into $k$-Graph system with the visualization of the computation steps.
Users can interact with the system using various parameters to enhance visualization and interpretation.

The GUI consists of five primary frames, as depicted in Figure~\ref{fig:frames}: the {\it Clustering Comparison Frame} (Figure~\ref{fig:frames}(1.1)), the {\it Benchmark Frame} (Figure~\ref{fig:frames}(1.2)), the {\it Graph Frame} (Figure~\ref{fig:frames}(2)), the {\it Interpretability Test Frame} (Figure~\ref{fig:frames}(3)), and the {\it Under the Hood Frame} (Figure~\ref{fig:frames}(4)).

\noindent \textbf{[Clustering comparison Frame]} This frame is divided into four sub-windows. 
The first three sub-windows display the time series of a user-selected dataset organized according to the clustering partition of $k$-Graph and two baselines (the ARI of each method is provided). 
The user can select individual time series and zoom to observe specific subsequences. 
The colors of the time series are based on the true labels of the datasets. 
Thus, the user can quickly assess the clustering partition quality (mixed colors mean low clustering accuracy).
Finally, the last sub-window corresponds to the time series organized according to the true labels.

\noindent \textbf{[Benchmark Frame]} This Frame depicts an overall accuracy evaluation of $k$-Graph against 14 baselines. The user can select the evaluation measure (among four measures) and filter the time series based on the dataset types, the time series length, the number of classes, and the number of time series. A box plot showing the performances of $k$-Graph and the 14 baselines is updated based on the filters.

\noindent \textbf{[Graph Frame]:}  This frame depicts the resulting graph embedding of the time series dataset obtained by $k$-Graph (Figure~\ref{fig:frames}(B)). It facilitates the exploration of nodes and edges.
For each node, users can visualize exclusivity and representativity values and the patterns they represent. The subsequences contained in the selected node are highlighted in the time series (below the graph). Additionally, users can filter nodes based on specific exclusivity and representativity values. The nodes and edges are colored if their representativity and exclusivity exceed the values the user selects.

\noindent \textbf{[Interpretability test Frame]:} This frame is a quiz. First, the user selects one dataset and a clustering method ($k$-Graph, $k$-Means, or $k$-Shape). Then, 5 time series (randomly picked in the selected dataset) will appear. The user's objective is to identify the cluster to which the time series has been assigned (by the selected clustering method). 
To assist the user, we provide the clusters' representation of the user-selected clustering method (the centroids for $k$-Means and $k$-Shape and the subgraph corresponding to the time series for $k$-Graph). 
After answering, a score appears. A high score means that the representation of clusters is highly interpretative.

\noindent \textbf{[Under the Hood Frame]} This frame enables users to explore the internal operations performed by $k$-Graph to obtain the final clusters (Figure~\ref{fig:frames}(C)). Users can visualize the values of the interpretability factor $W_e(\ell)$ and the consistency factor $W_c(\ell)$ computed for each generated length. Additionally, the frame provides information on the computation of the features matrix and the consensus matrix, which are crucial for obtaining the final clustering. All the plots in this frame are automatically updated according to the dataset selected by the user.

\section{Demonstration Scenarios}
This demo has three goals: 
(i) to propose an interactive application for comparing time series clustering methods;
(ii) to enable users to interpret the clustering of a time series dataset using $k$-Graph and evaluate its interpretability power; and (iii) to allow users to explore the steps of $k$-Graph. 

\noindent \textbf{[Scenario 1: Interpretability Test]} 
In this scenario, users are directed to the interpretability test frame (Figure~\ref{fig:frames}(3)). They will select a dataset and a clustering method using the drop-down menus in the sidebar. Users are initially asked to choose either $k$-Means or $k$-Shape. Then, we will ask the user to answer five questions (randomly generated) with one time series each (from the selected dataset). In these questions, the objective is to find which cluster the time series has been assigned to based on the selected clustering approach. At first, the user will only have the possibility to display the centroid of each cluster. After answering the questions, a score will appear. We then ask the user to select $k$-Graph and to redo the quiz. This time, the user can use the Graph obtained with $k$-Graph. We will then compare the user's scores obtained with the $k$-graph and the other baseline.

\noindent \textbf{[Scenario 2: Exploring $k$-Graph]} 
After completing the first scenario, users can explore the graph representation obtained with $k$-Graph. Thus, this scenario starts in the graph frame (Figure~\ref{fig:frames}(2)). 
We ask the user to identify the discriminant features of each cluster identified by $k$-Graph.
The user can select a node and visualize which patterns it represents, as well as a histogram for the exclusivity and representativity of each cluster for that node (bottom right of Figure~\ref{fig:frames}(B)). 
The nodes and edges are automatically colored based on predefined threshold values for representativity, $\lambda$, and exclusivity, $\gamma$. 
The user can change these values using the advanced setting window (top right of Figure~\ref{fig:frames}(B)).
This allows the user to filter the graph according to $\lambda$ and $\gamma$. 
The objective is to find the correct value of $\gamma$ and $\lambda$ so we have at least one colored node per cluster.
Then, the user can verify if the identified patterns are consistent with the true cluster labels
(in Figure~\ref{fig:frames}(1.1)).

\noindent \textbf{[Scenario 3: Under the Hood]} 
In the final scenario, users are directed to the "Under the Hood" frame (Figure~\ref{fig:frames}(4)). 
In this frame, the user can dive into the different algorithmic steps of $k$-Graph, and try to answer some key questions, such as {\it How is the subsequence length selected for the graph displayed in the Graph frame?}, or {\it How is the graph used to cluster the time series?},
We will ask the user to answer based on their observations from the two previous scenarios. 
The user can then reveal the information needed to produce the answers and view the corresponding visualizations (specific to the selected dataset). 
Thus, the user can change datasets to enhance their understanding of our method.

\section{Conclusions}
We demonstrate Graphint, a system that enables users to understand and interpret the time series clustering performed by $k$-Graph. 
Graphint represents time series as a graph, visualizes this graph, and uses the graph to identify subsequences that can discriminate among clusters. 
Our system allows users to dive into the algorithmic steps of $k$-Graph, and observe their operation on different datasets.

\subsubsection*{Acknowledgment} Work partially funded by EU Horizon projects AI4Europe (101070000),
TwinODIS (101160009), ARMADA (101168951), DataGEMS (101188416) and
RECITALS (101168490).

\bibliographystyle{IEEEtran}
\bibliography{sample}

\end{document}